\documentclass[journal]{IEEEtran}

\usepackage{epsfig}
\usepackage{graphicx}
\usepackage{amsmath}
\usepackage{amssymb}
\usepackage{enumitem}
\usepackage{bm}
\usepackage{booktabs}
\usepackage{subcaption}
\usepackage{multirow}

\usepackage{url}

\usepackage{color}
\usepackage{amsmath}
\usepackage{amssymb}
\usepackage{bbding}
\usepackage{bm}
\usepackage{booktabs}
\usepackage{wrapfig}
\usepackage{makecell}

\usepackage[colorlinks,
            linkcolor=red,
            anchorcolor=blue,
            citecolor=green
            ]{hyperref}

\def\eg{\emph{e.g.}}
\def\ie{\emph{i.e.}}

\def\vs{\emph{vs}}

\makeatletter
\newcommand*\bigcdot{\mathpalette\bigcdot@{.7}}
\newcommand*\bigcdot@[2]{\mathbin{\vcenter{\hbox{\scalebox{#2}{$\m@th#1\bullet$}}}}}
\makeatother

\usepackage{mathtools}

\usepackage{tabularx} % in the preamble
\usepackage{adjustbox}

\ifCLASSOPTIONcompsoc
  \usepackage[nocompress]{cite}
\else
  \usepackage{cite}
\fi

\begin{document}
\title{Region Aware Video Object Segmentation\\ with Deep Motion Modeling}

\author{Bo Miao, 
Mohammed Bennamoun,~\IEEEmembership{Senior Member,~IEEE,}\\
Yongsheng Gao,~\IEEEmembership{Senior Member,~IEEE,} 
Ajmal Mian,~\IEEEmembership{Senior Member,~IEEE}% <-this % stops a space
\IEEEcompsocitemizethanks{\IEEEcompsocthanksitem B. Miao, M. Bennamoun, and A. Mian are with the Department of Computer Science and Software Engineering, The University of Western Australia, Perth, Crawley, WA 6009, Australia (e-mail: bo.miao, mohammed.bennamoun, ajmal.mian@uwa.edu.au). %\protect\\
\IEEEcompsocthanksitem Y. Gao is with the School of Engineering, Griffith University, Brisbane, QLD 4111, Australia (e-mail: yongsheng.gao@griffith.edu.au). %\protect\\
}% <-this % stops an unwanted space
% \thanks{Manuscript received December 20, 2020, modified November 23, 2021.}
\thanks{This research was funded by the Australian Research Council Industrial Transformation Research Hub IH180100002. Professor Ajmal Mian is the recipient of an Australian Research Council Future Fellowship Award (project number FT210100268) funded by the Australian Government.}
}


\markboth{ }%
{Hou \MakeLowercase{\textit{et al.}}: Adaptive Affinity for Associations in Multi-Target Multi-Camera Tracking}


\maketitle

\begin{abstract}
Current semi-supervised video object segmentation (VOS) methods usually leverage the entire features of one frame to predict object masks and update memory. This introduces significant redundant computations. To reduce redundancy, we present a Region Aware Video Object Segmentation (RAVOS) approach that predicts regions of interest (ROIs) for efficient object segmentation and memory storage. RAVOS includes a \emph{fast} object motion tracker to predict their ROIs in the next frame. For efficient segmentation, object features are extracted according to the ROIs, and an object decoder is designed for object-level segmentation. For efficient memory storage, we propose motion path memory to filter out redundant context by memorizing the features within the motion path of objects between two frames. Besides RAVOS, we also propose a large-scale dataset, dubbed OVOS, to benchmark the performance of VOS models under occlusions. Evaluation on DAVIS and YouTube-VOS benchmarks and our new OVOS dataset show that our method achieves state-of-the-art performance with significantly faster inference time, \eg, 86.1 $\mathcal{J}$\&$\mathcal{F}$ at 42 FPS on DAVIS and 84.4 $\mathcal{J}$\&$\mathcal{F}$ at 23 FPS on YouTube-VOS.
\end{abstract}


% \fi
\IEEEpeerreviewmaketitle

\section{Introduction}
\IEEEPARstart{V}{ideo} object segmentation (VOS) is a fundamental research topic in visual understanding, with the aim to segment target objects in video sequences. 
VOS enables machines to sense the motion pattern, location, and boundaries of the objects of interest in videos~\cite{MATNet}, which can foster a wide range of applications, \eg, augmented reality, video editing, and robotics. This work focuses on semi-supervised VOS, where object segmentations given on the first-frame are leveraged to segment and track objects in subsequent frames. A practical semi-supervised VOS method should be able to segment the objects of interest efficiently and accurately under challenging scenarios, such as occlusions, large deformations, similar appearances, background confusion, and scale variations.

Recent semi-supervised VOS methods mainly follow one of two paradigms: detection-based~\cite{OSVOS,OnAVOS,E_OSVOS} and memory-based~\cite{STM,CFBI,AOT,STCN}. Detection-based methods usually rely on online adaption to make the model object-specific, while memory-based methods adopt memory networks to memorize and propagate spatio-temporal features across frames for object segmentation. Methods in the latter paradigm have recently drawn significant research attention due to their exceptional accuracy. These methods either perform non-local matching~\cite{STM,STCN} or local-matching~\cite{MAST,MAMP} for mask propagation.

Although current memory-based methods have shown promising performance, memorizing and segmenting the entire features of one frame inevitably introduces redundant computations and slows down the process. Some methods have attempted to accelerate VOS by introducing additional instance segmentation or detection networks~\cite{DMN-AOA,TAN_DTTM}, template matching modules~\cite{SiamMask,SAT,FAVOS}, or optical flow~\cite{RMNet} to create regions of interest (ROIs) and then performing local segmentation. However, these local segmentation methods are either not accurate enough or still time-consuming given the additional computational overhead. Therefore, developing an effective method that avoids redundant computations and memory storage, while maintaining high segmentation accuracy is significant for improving the overall semi-supervised VOS performance.

\begin{figure}[t]
\centering
\includegraphics[width=0.95\columnwidth]{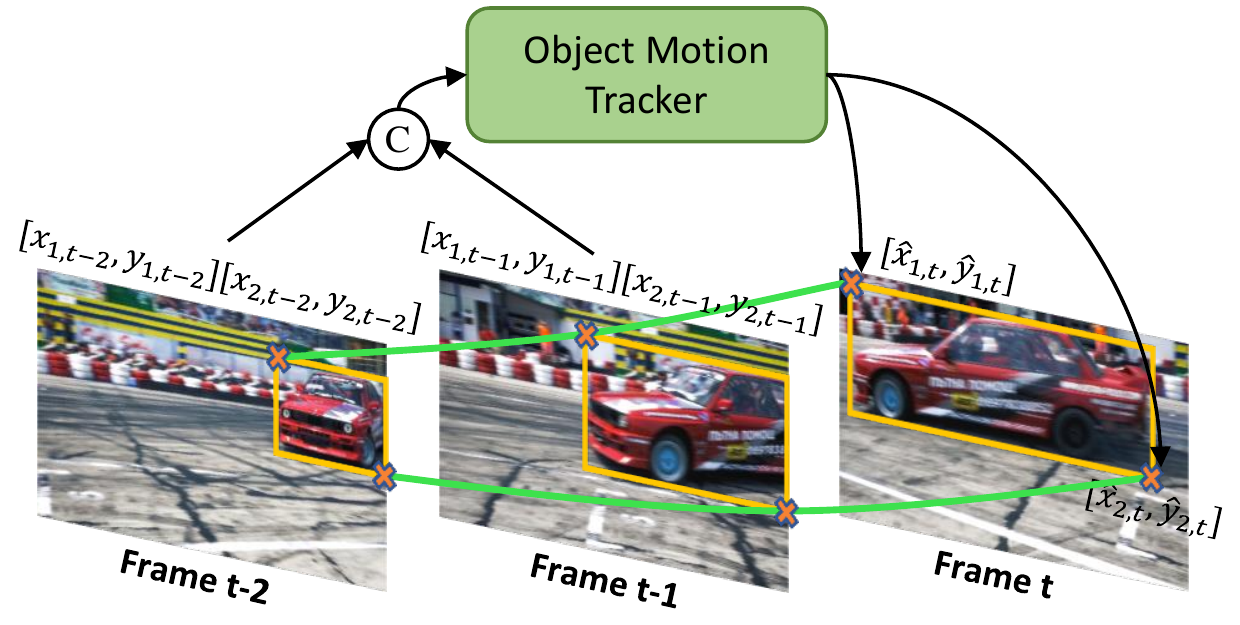}
\caption{Concept of our object motion tracker. Each object is tracked across frames by predicting the parameters of motion functions using the position features of the object in the previous frames.}
\label{fig:tracker}
\end{figure}

In this paper, we propose a novel Region Aware Video Object Segmentation (RAVOS) approach, which enables multi-object tracking and ROI prediction to achieve fast and accurate semi-supervised VOS with less memory burden. First, a lightweight object motion tracker (OMT) is proposed to estimate the parameters of motion functions using the position information of instances in past frames for object tracking and ROI prediction, as shown in Fig.~\ref{fig:tracker}. Since the position features, rather than costly image features, are used for tracking, OMT achieves about 5000 FPS on a single GPU. To enable efficient object segmentation, we extract object features based on the predicted ROIs and adopt a designed object decoder that uses object skip connections for object-level segmentation. Second, we propose motion path memory (MPM) to filter out redundant context by memorizing the features within the motion path of objects between two frames. Hence, redundant segmentation and memory storage are alleviated significantly. 

Occlusion is a challenging scenario for matching-based VOS methods due to the similar appearance and position of objects. Currently, no large-scale datasets are designed to evaluate semi-supervised VOS models under occlusions specifically. To fill this gap, we create a large-scale occluded video object segmentation dataset, coined OVOS, based on the OVIS dataset~\cite{OVIS}. We further evaluate our method and the state-of-the-art STCN~\cite{STCN} on the OVOS dataset to verify their ability in occlusion scenarios.

We perform extensive experiments on benchmark datasets, \ie, DAVIS and YouTube-VOS, and our new OVOS dataset to evaluate the performance of our method. RAVOS achieves state-of-the-art overall performance compared to existing methods. For instance, it achieves 86.1 $\mathcal{J}$\&$\mathcal{F}$ with 42 FPS on DAVIS 2017 validation set, outperforming current methods in both accuracy and inference speed. Our main contributions are summarized as follows:
\begin{itemize}
\item[$\bullet$] We propose motion path memory (MPM), which memorizes the features within the motion path of objects to mitigate redundant memory storage and to accelerate feature matching and propagation.
\item[$\bullet$] We propose a fast (5000 FPS) object motion tracker to track objects across frames by predicting the parameters of motion functions. This enables object-level segmentation with the help of our designed object decoder, which leverages object skip connections.
\item[$\bullet$] We create an occluded video object segmentation (OVOS) dataset and compare the performance of RAVOS with an existing method on it. To the best of our knowledge, this is the first time a semi-supervised VOS method is evaluated on a large-scale dataset with occlusions. The dataset is available at \url{http://ieee-dataport.org/9608}.
\item[$\bullet$]
Experiments on DAVIS, YouTube-VOS, and OVOS datasets show that our method achieves state-of-the-art performance while running twice as fast as existing ones.
\end{itemize}

\section{Related Work}\label{related work}
\textbf{Semi-supervised Video Object Segmentation.} Semi-supervised VOS aims to leverage the ground truth object masks provided (only) for the first frame to segment the entire video sequence at pixel level. Before the rise of deep learning, traditional methods usually adopted graphical models~\cite{TraditionalGraphVOS} or optical flow~\cite{TraditionalFlowVOS} for video segmentation. Recent studies of semi-supervised VOS mainly focus on deep neural networks because of their unmatched performance.

Early deep learning-based methods often fine-tune the networks on each video sequence during inference, making them focus on different target objects~\cite{MaskTrack,MoNet,MHPVOS,TAN_DTTM,FRTM}. For example, OSVOS~\cite{OSVOS} and its variants~\cite{OnAVOS,OSVOS_S,E_OSVOS} fine-tune their networks on the first frame or confident middle frames. Lucid Tracker~\cite{Lucid} and PReMVOS~\cite{PReMVOS} use data augmentation to generate plenty of synthetic frames for online fine-tuning. Despite their satisfying results, online fine-tuning severely limits the inference speed of networks and leads to over-fitting. To accelerate VOS, DMN-AOA~\cite{DMN-AOA} adopts instance segmentation network to generate plenty of ROIs and then perform local segmentation after non-maximum suppression operation. SAT~\cite{SAT} incorporates template matching for object localization.

To achieve higher segmentation accuracy, recent works aim to leverage spatio-temporal feature propagation~\cite{PLM,RGMP,FSNet,G-FRTM,CapsuleVOS} or pixel-wise feature matching~\cite{PML,RaNet,VideoMatch,TVOS,SSTVOS,CFBI,LCM} to guide VOS. The former propagates spatio-temporal features implicitly across frames. Among them, RVOS~\cite{RVOS} and DyeNet~\cite{DyeNet} adopt recurrent neural networks to propagate spatio-temporal features. AGAME~\cite{A_GAME} proposes a fusion module that integrates spatio-temporal features with the appearance features of the current frame. 
The latter computes spatio-temporal correspondences for mask propagation.
PML~\cite{PML} adopts pixel-wise metric learning and classifies pixels based on a nearest-neighbor method.
STM~\cite{STM} and its variants~\cite{AFB_URR,GC} memorize spatio-temporal features and perform non-local spatio-temporal matching for temporal association. To enable unsupervised training, MAST~\cite{MAST} and MAMP~\cite{MAMP} use a self-supervised photometric reconstruction task to learn to construct spatio-temporal correspondences without any mask annotations. To accelerate association, RMNet~\cite{RMNet} leverages optical flow to perform regional matching. Based on the observation that the dot product affinity leads to poor memory usage, STCN~\cite{STCN} adopts L2 similarity for affinity measurement. 

The above methods achieve good performance on semi-supervised VOS. However, they either require to segment and memorize the entire features of one frame, which leads to redundant computations and memory storage, or rely on extra time-consuming networks, \eg, optical flow, to locate ROIs. These problems restrict the deployment of VOS in memory-constrained real-time applications. Hence, an effective ROI localization and segmentation method is needed for fast and accurate VOS. We propose RAVOS, which contains an extremely fast object motion tracker to predict ROIs and leverages object-level segmentation and motion path memory for efficient segmentation and memorization.

\begin{figure*}[t]
\centering
\includegraphics[width=0.95\linewidth]{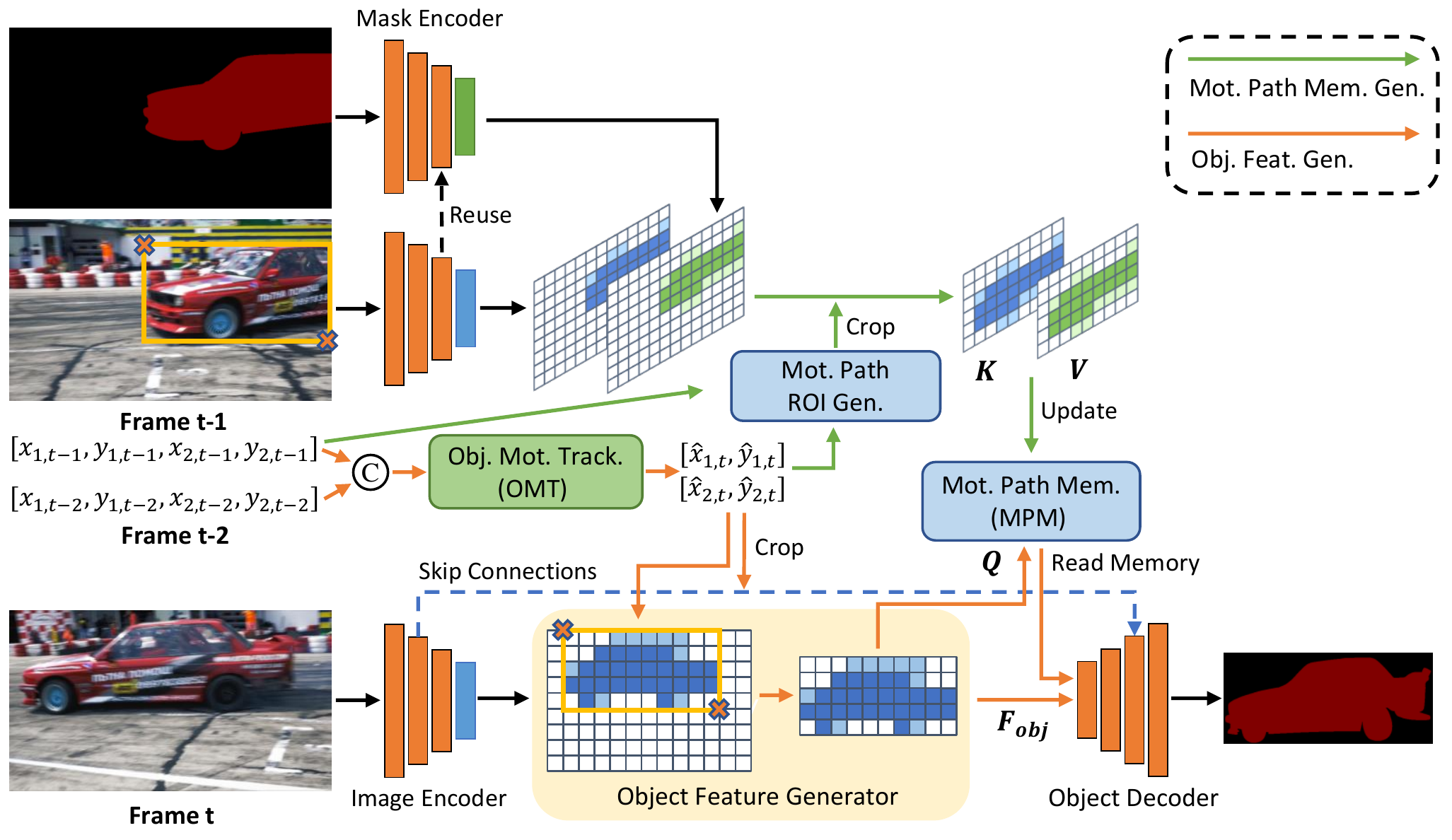}
\caption{RAVOS architecture. $Q$, $K$, $V$, and $F$ denote the $query$ of frame $t$, $key$ of the memory, $value$ of the memory, and appearance features of frame $t$, respectively. The proposed OMT estimates the ROIs of target objects. Each object is then decoded and segmented at object level according to the object ROIs, MPM, and object decoder.} \label{fig:framework}
\end{figure*}

\textbf{Multi-Object Tracking.} Multi-object tracking (MOT) aims to continuously estimate the trajectories of target objects across frames. Object detection, association, and motion estimation are three key components of MOT. Among them, CenterTrack~\cite{CenterTrack} adopts a detection network to detect object centers and predict motion offsets for tracking. TraDeS~\cite{TraDeS} estimates motion offsets to track objects, and combines the tracking results with detection results for MOT. DMMNet~\cite{DMMNet} leverages spatio-temporal features to predict tracklets for tracking. TT17~\cite{TT17} proposes an iterative clustering method to generate multiple high confidence tracklets for objects. ByteTrack~\cite{ByteTrack} incorporates low-confident boxes for association to dig out objects. DAN~\cite{DAN} proposes an affinity refinement module for more comprehensive associations.

With the help of object trackers, we can locate ROIs for region aware segmentation and memorization. However, directly using existing MOT methods for tracking in VOS will introduce redundant architectures and computations, violating the lightweight and real-time VOS performance requirements. In this work, we propose OMT to meet the lightweight and real-time processing requirements. Instead of using image features for tracking, OMT leverages the object position information in previous frames to predict the parameters of motion functions.

\textbf{Memory Networks.} Memory networks aim to capture the long-term dependencies by storing temporal features or different categories of features in a memory module. LSTM \cite{LSTM} and GRU \cite{GRU} implicitly represent spatio-temporal features with local memory cells in a highly compressed way limiting the representation ability. Memory networks \cite{MemNet} were introduced to explicitly store the important features. A classical memory network-based VOS method is STM \cite{STM} which incrementally adds uncompressed features of past frames to the memory bank, and performs non-local matching to propagate spatio-temporal features. However, the background features are highly redundant. In this work, we introduce motion path memory which filters redundant context (background far from objects) while still keeping important context (foreground and nearby background).

\section{Method}
We propose RAVOS, an efficient and accurate semi-supervised VOS method as shown in Fig.~\ref{fig:framework}. In a nutshell, RAVOS is developed based on matching-based VOS framework and contains five parts: \emph{feature extraction}, \emph{ROI prediction}, \emph{memory storage}, \emph{memory propagation}, and \emph{object segmentation}.

RAVOS adopts ResNet-50 and ResNet-18~\cite{ResNet} to encode image features (\textit{key} and \textit{query}) and mask features (\textit{value}), separately. After extracting features via encoders, RAVOS uses the proposed OMT to track objects and predict their ROIs in frame $t$, \ie, $\hat{R}_{t} \in [\hat{x}_{1,t},\hat{y}_{1,t},\hat{x}_{2,t},\hat{y}_{2,t}]$. Next, $R_{t-1}$ and the predicted object ROIs $\hat{R}_{t}$ are forwarded to the motion path ROI generator to generate memory ROIs in frame $t$-1, and MPM is updated by the features within memory ROIs. Then, object-level features of frame $t$ are extracted according to the predicted object ROIs $\hat{R}_{t}$ and their corresponding spatio-temporal features are retrieved from the memory bank. Finally, an object decoder with object skip connections is used to segment each object.

\begin{figure*}[t]
\centering
\includegraphics[width=0.97\linewidth]{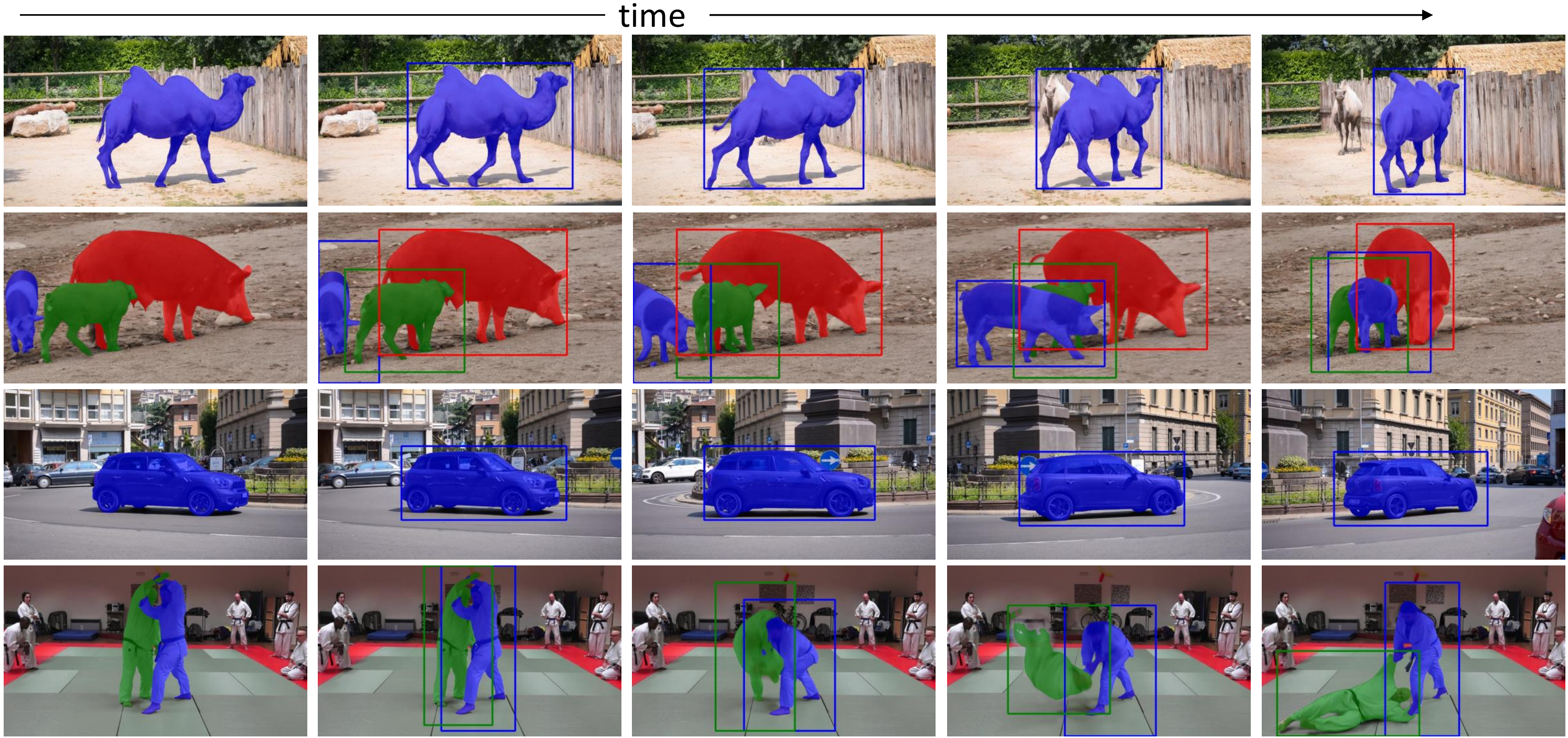}
\caption{Visualization of the tracking results predicted by OMT. The first column denotes the reference frames for mask propagation, and the segmentation is performed within the object ROIs.} 
\label{fig:trackerresults}
\end{figure*}

\begin{figure}[t]
\centering
\includegraphics[width=0.95\columnwidth]{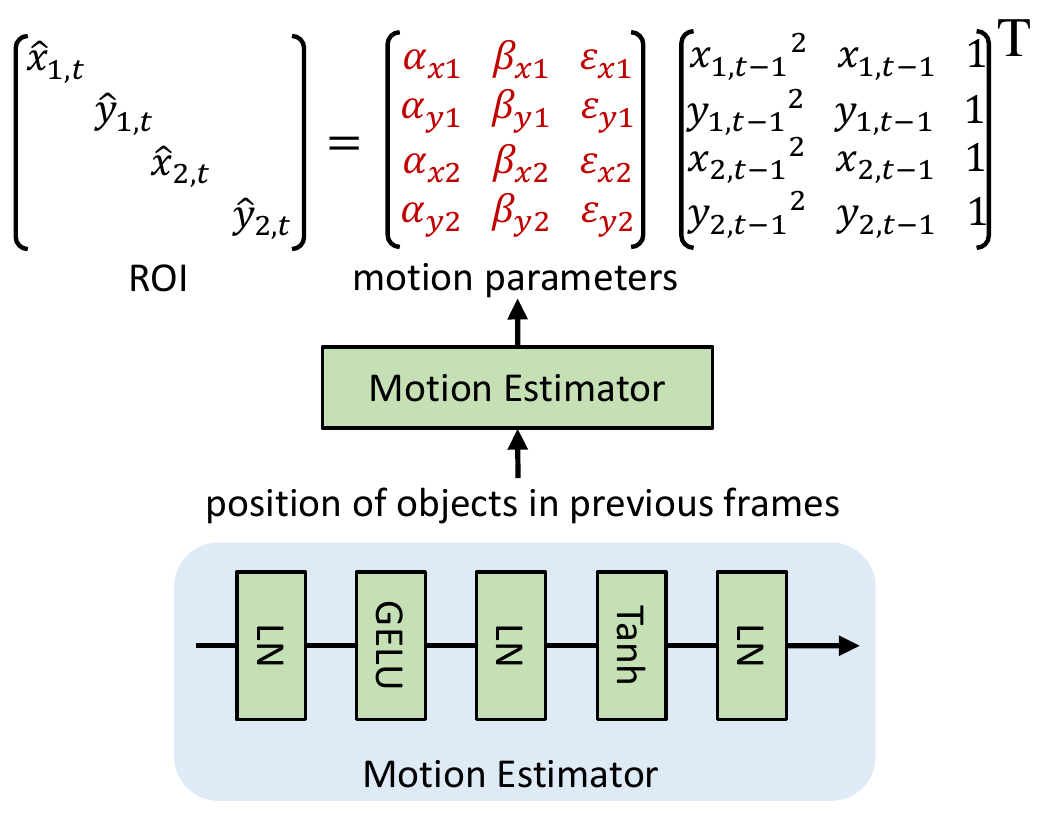}
\caption{Object motion tracker. The proposed tracker encodes the position of each object in previous frames into the parameters of quadratic motion functions for tracking.}
\label{fig:trackerarchitecture}
\end{figure}

\subsection{Feature Extraction}
Following previous works~\cite{STCN,STM}, we adopt ResNet-50 and ResNet-18 (excluding their last stages) \cite{ResNet} as the \emph{image} and \emph{mask} encoders, respectively. Inputs are downsampled by 1/16 via encoders. For the \emph{image} encoder, one additional 3$\times$3 convolutional layer is used on top of the $res4$ features to extract $key$ $K \in{\mathbb{R}^{HW \times 64}}$ or $query$ $Q \in{\mathbb{R}^{HW \times 64}}$ for matching, and another 3$\times$3 convolutional layer is leveraged on top of the $res4$ features to compute appearance features $F \in{\mathbb{R}^{HW \times 512}}$ to assist object segmentation. Image features at the middle layers are also saved to extract object skip connections for the object decoder. For the \emph{mask} encoder, two residual blocks as well as one CBAM block \cite{CBAM} are used on top of the $res4$ features to extract $value$ $V \in{\mathbb{R}^{HW \times 512}}$ for each object.

\subsection{Object Motion Tracker}
\label{Tracker}

An effective tracker is imperative for ROI prediction and efficient regional semi-supervised VOS. Existing deep learning-based MOT methods use appearance features for tracking. Although appearance features lead to good MOT performance, directly incorporating such techniques into VOS is difficult to cater for the lightweight and real-time processing requirements. 

To address the problem, we propose a novel object motion tracker (OMT) that leverages object position features in previous frames to predict the parameters of instantaneous motion functions for MOT, as shown in Fig.~\ref{fig:trackerarchitecture}. Specifically, given the normalized position information of the $i$th object in the previous frames, \eg, $R_{t-2}^{i} \in [x_{1,t-2}^{i},y_{1,t-2}^{i},x_{2,t-2}^{i},y_{2,t-2}^{i}]$ and $R_{t-1}^{i} \in [x_{1,t-1}^{i},y_{1,t-1}^{i},x_{2,t-1}^{i},y_{2,t-1}^{i}]$. Where $[x_{1},y_{1}]$ denotes the top-left and $[x_{2},y_{2}]$ denotes the bottom right corner. OMT uses a deep motion estimator to aggregate the position features of the object in previous frames and to predict the parameters of motion functions for each corner. In this work, we choose the simple yet effective quadratic function as the motion function template. Next, the object ROI $\hat{R}_{t}^{i}$ in the current frame is predicted by plugging $R_{t-1}^{i}$ into the estimated motion functions:
\begin{equation}
\begin{split}
    \hat{x}_{1,t} = \alpha_{x1} {x_{1,t-1}}^{2} + \beta_{x1} x_{1,t-1} + {\epsilon}_{x1} - \phi \\
    \hat{y}_{1,t} = {\alpha}_{y1} {y_{1,t-1}}^{2} + \beta_{y1} y_{1,t-1} + {\epsilon}_{y1} - \phi \\
    \hat{x}_{2,t} = {\alpha}_{x2} {x_{2,t-1}}^{2} + \beta_{x2} {x}_{2,t-1} + {\epsilon}_{x2} + \phi \\
    \hat{y}_{2,t} = {\alpha}_{y2} {y_{2,t-1}}^{2} + \beta_{y2} {y}_{2,t-1} + {\epsilon}_{y2} + \phi
\end{split}
\label{eq1}
\end{equation}
where $\alpha$, $\beta$, and $\epsilon$ are the predicted parameters of motion functions, and $\phi$ denotes the padding of bounding boxes. Finally, the segmentation operates within the object ROI $\hat{R}_{t}^{i}$ to reduce computation and update the ROI. 
% Version 1
Different from previous template matching-based methods~\cite{SAT,FAVOS}, OMT does not rely on feature matching that struggles to handle objects with similar appearances. 
Unlike optical flow-based methods~\cite{RMNet}, OMT leverages the lightweight but crucial position features rather than costly appearance features to estimate ROIs. The lightweight framework of OMT enables it to perform at 5000 FPS on a single GPU which is about \emph{100}$\times$ faster than the prevalent RAFT optical flow \cite{RAFT}.
Moreover, OMT predicts motion functions to generate a definite ROI for each object rather than generating many proposals using additional detection networks like~\cite{DMN-AOA,TAN_DTTM} making the non-maximum suppression (NMS) operation redundant.
The visualization of the tracking results shows that our efficient OMT can generate sufficiently accurate ROIs for regional VOS (see Fig.~\ref{fig:trackerresults}).

\begin{figure}[t]
\centering
\includegraphics[width=\columnwidth]{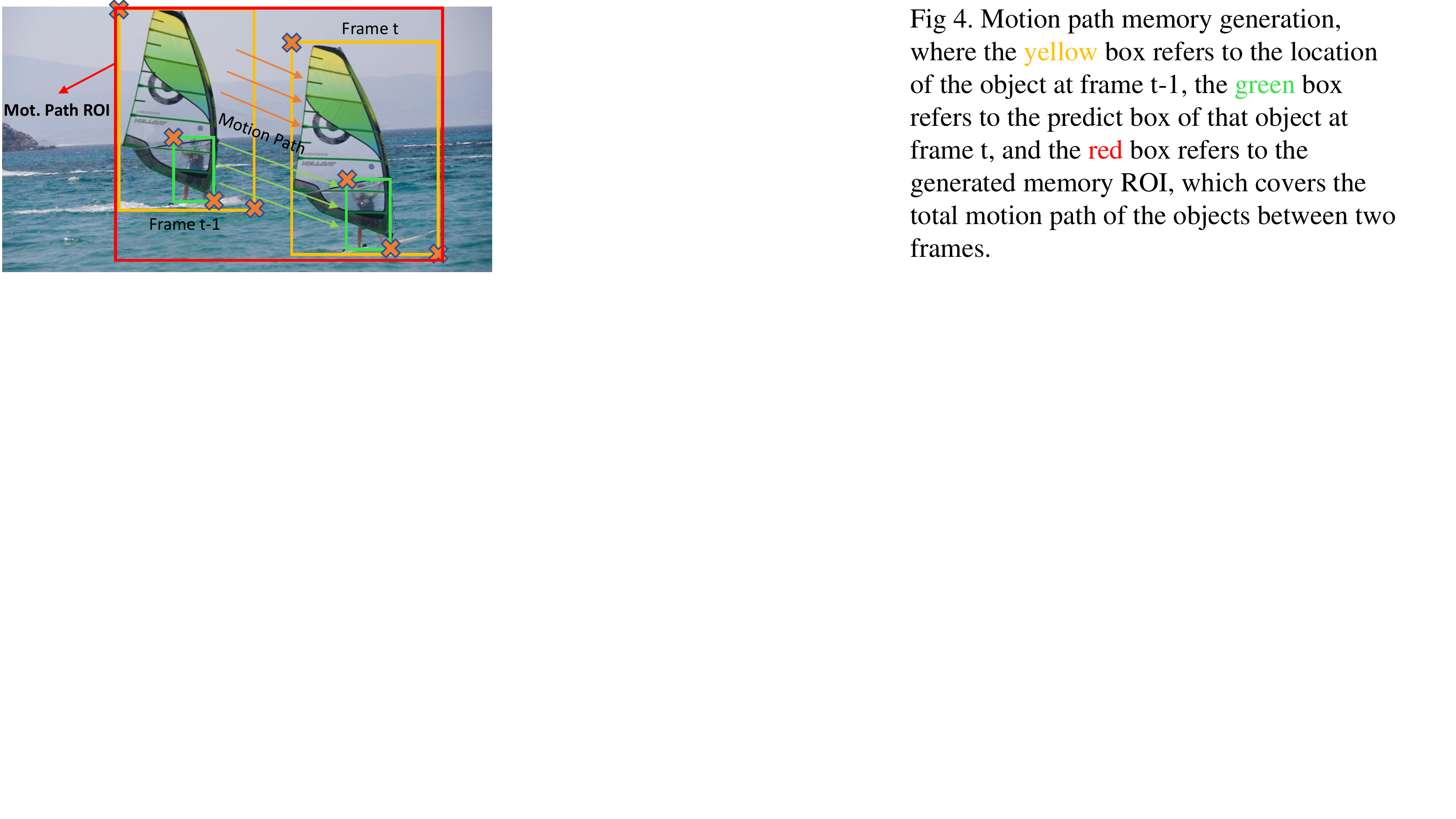}
\caption{Motion path ROI generation. The yellow and green bounding boxes denote the object ROIs of the sailboat and person in two frames, respectively.
The red bounding box refers to the generated motion path ROI for frame $t$-1, which covers the motion path of all objects between two frames and filters redundant background far from objects. All bounding boxes in frame $t$ are predicted by OMT.}
\label{fig:memory}
\end{figure}

\subsection{Motion Path Memory}
Previous methods~\cite{STCN,MAMP} have shown that only a few positions in the memory are helpful for the association of a query point. Moreover, the proposed RAVOS only segments the ROIs generated by OMT. Therefore, memorizing the entire features of one frame will include redundant background context, which is far from foreground objects.
The redundant context impedes the deployment of efficient VOS. 

For redundancy reduction, we present motion path memory (MPM) to memorize the critical context, \ie, the foreground and nearby background. As shown in Fig.~\ref{fig:memory}, MPM generates the motion path of each object between two frames and then memorizes features within the united motion paths to filter redundant context. Specifically, before memorizing the features of frame $t$-1, we first predict the ROIs of objects $\hat{R}_{t} = \{\forall \hat{R}_{t}^{i}, i \in [1, ..., N] \}$ in the next frame $t$ using OMT. The motion path of each object between two frames is then extracted according to the position of the object in the two frames. Finally, the union of all motion paths is created as the memory ROIs:
\begin{equation}
ROI = U(\{\forall U(R_{t-1}^{i}, \hat{R}_{t}^{i}), i \in [1, ..., N] \})
\label{eq2}
\end{equation}
where $U$ denotes the union operation and $N$ is the number of objects. In that case, redundant background features outside the memory ROI will not be used to update the memory. Therefore, the proposed MPM not only accelerates feature matching and propagation, but also reduces the memory footprint for efficient deployment.

\subsection{Memory Propagation}
As in common matching-based VOS methods~\cite{STM,STCN}, the memory module propagates mask features across frames for segmentation according to the pixel-wise affinity between all query and memory pixels. In this work, we perform regional matching using L2 similarity to compute the affinity. 

To illustrate, we first define $Q \in{\mathbb{R}^{HW \times 64}}$, $K \in{\mathbb{R}^{N \times 64}}$, $V \in{\mathbb{R}^{N \times 512}}$ as the $query$ of the current frame, $key$ of the memory, and $value$ of the memory, respectively. Where $H$ and $W$ are the feature height and width, and $N\ll THW$ denotes the number of positions in the memory. For each object, after predicting the object ROI $\hat{R_{t}}$ at frame $t$, we crop $query$ to obtain $object$ $query$ $Q_{obj} \in{\mathbb{R}^{S1S2 \times 64}}$, where $S1$ and $S2$ denote the height and width of $\hat{R_{t}}$ at feature scale and $S1S2\ll HW$. Then, the affinity between $Q_{obj}$ and $K$ is computed as:
\begin{equation}
W^{i,j} = \frac{exp(\langle Q_{obj}^{i},K^{j} \rangle)}{\sum_{j} exp(\langle Q_{obj}^{i},K^{j} \rangle)}
\label{eq3}
\end{equation}
where $i$ and $j$ are the locations in $Q_{obj}$ and $K$, respectively. $\langle \cdot , \cdot \rangle$ denotes the L2 similarity between two vectors. Finally, each query position retrieves spatio-temporal features from the memory based on the computed affinity:
\begin{equation}
V_{st}^{i} = \sum_{j} W^{i,j}V^{j}
\label{eq4}
\end{equation}

\begin{figure}[t]
\centering
\includegraphics[width=\columnwidth]{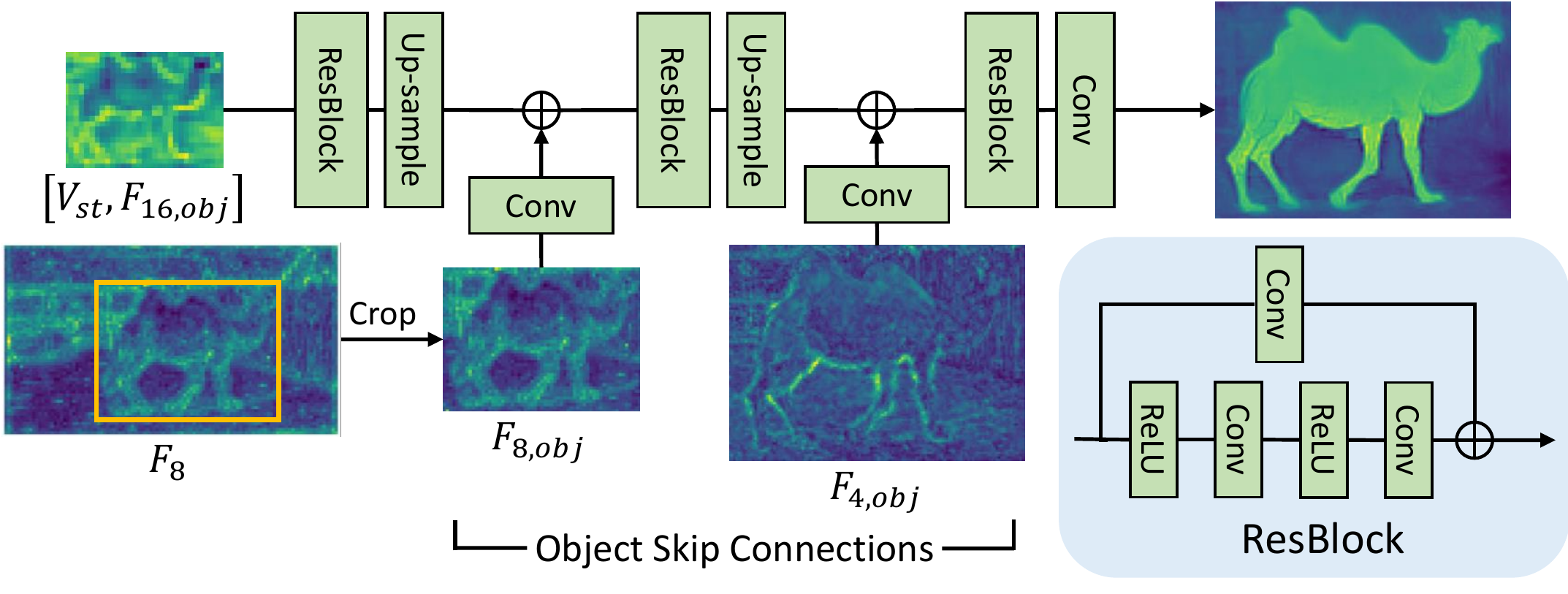}
\caption{Object decoder. $F_{16,obj}$ and $V_{st}$ denote the appearance features of current frame and the queried spatio-temporal features, respectively. $F_{8,obj}$ and $F_{4,obj}$ are object skip connections extracted from the middle layer features of the image encoder.} 
\label{fig:decoder}
\end{figure}

\begin{figure*}[t]
\centering
\includegraphics[width=0.97\linewidth]{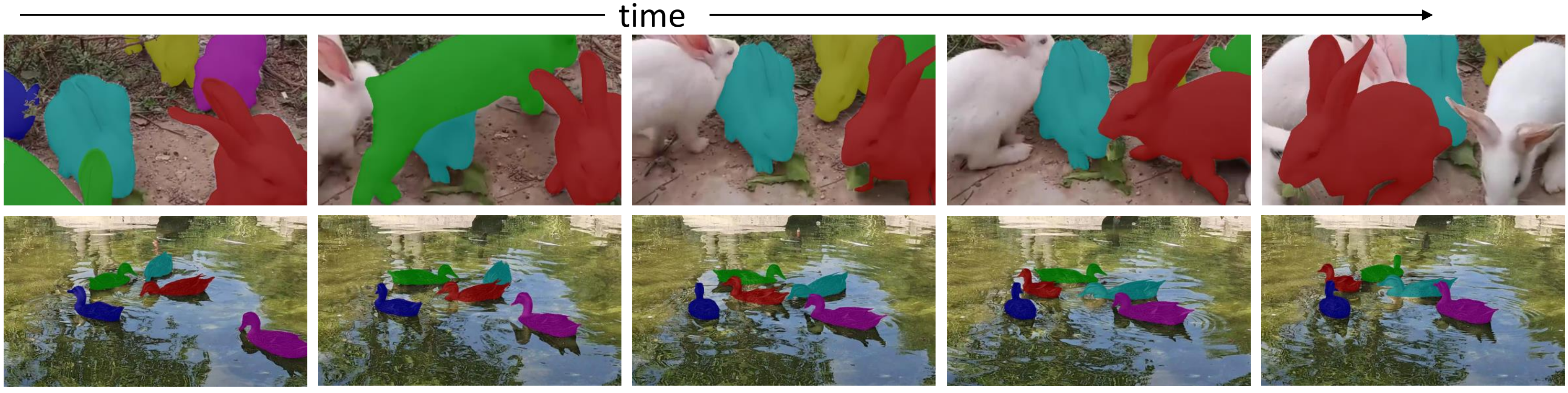}
\caption{Examples of video sequences in OVOS dataset.} 
\label{fig:ovos}
\end{figure*}

\subsection{Object-level Segmentation}
To reduce redundant computations without losing important information, we predict object ROIs and decode object features. As shown in Fig.~\ref{fig:decoder}, residual blocks with object skip connections are leveraged to build the object decoder. Specifically, the object decoder concatenates the object appearance features of the current frame $F_{obj}$ and their corresponding spatio-temporal features $V_{st}$ obtained from the memory as input, and progressively increases the object feature resolution from 1/16 to 1/4. The decoded object features are used to predict object probability maps and are up-sampled to the input resolution. Finally, the object probability maps are projected on the image probability map, and Argmax is used for segmentation.

\begin{table*}[t]
\centering
\caption{Quantitative evaluation on DAVIS 2016 and 2017 validation sets. $RS$: regional segmentation. $^*$: re-timed on our machine for fair-comparison.} \label{tab:davis}
\begin{tabular}{lccccccccc}
\toprule 
& & \multicolumn{4}{c}{DAVIS 2017} & \multicolumn{4}{c}{DAVIS 2016} \\
\midrule 
Method & RS & \( \mathcal{J} \)\&\( \mathcal{F} \) & \( \mathcal{J} \) & \( \mathcal{F} \) & FPS & \( \mathcal{J} \)\&\( \mathcal{F} \) & \( \mathcal{J} \) & \( \mathcal{F} \) & FPS \\
\midrule 
OSVOS~\cite{OSVOS} & $\times$ & 60.3 & 56.6 & 63.9 & \textless1 & 80.2  & 79.8  & 80.6 & \textless1 \\
RGMP~\cite{RGMP} & $\times$ & 66.7 & 64.8 & 68.6 & \textless7.7 & 81.8  & 81.5  & 82.0 & 7.7 \\
FEELVOS~\cite{FEELVOS} & $\times$ & 71.5 & 69.1 & 74.0 & 2.0 &  81.7  & 81.1  & 82.2 & 2.2 \\
GC~\cite{GC} & $\times$ & 71.4 & 69.3 & 73.5 & \textless25 & 86.6 & 87.6 & 85.7 & 25\\
AFB-URR~\cite{AFB_URR} & $\times$ & 74.6 & 73.0 & 76.1 & 4 & - & - & - & - \\
KMN~\cite{KMN} & $\times$ & 82.8 & 80.0 & 85.6 & \textless8.3 & 90.5  & 89.5  & 91.5 & 8.3 \\
CFBI+~\cite{CFBI+} & $\times$ & 82.9 & 80.1 & 85.7 & 5.6 & 89.9 & 88.7  & 91.1 & 5.9 \\
SwiftNet~\cite{SwiftNet} & $\times$ & 81.1 & 78.3 & 83.9 & \textless25 & 90.4  & 90.5  & 90.3 & 25 \\
ASRF~\cite{ASRF} & $\times$ & 83.2 & 80.3 & 86.1 & - & 90.9  & 90.1  & 91.7 & - \\
LCM~\cite{LCM} & $\times$ & 83.5 & 80.5 & 86.5 & \textless8.5 & 90.7  & 89.9  & 91.4 & 8.5 \\
JOINT~\cite{JOINT} & $\times$ & 83.5 & 80.8 & 86.2 & 4 & - & - & - & - \\
HMMN~\cite{HMMN} & $\times$ & 84.7 & 81.9 & 87.5 & \textless10 & 90.8 & 89.6 & 92.0 & 10 \\
STM~\cite{STM} & $\times$ & 81.8 & 79.2 & 84.3 & 19$^*$ & 89.3  & 88.7 & 89.9 & 23$^*$ \\
R50-AOT~\cite{AOT} & $\times$ & 84.9 & 82.3 & 87.5 & 24$^*$ & 91.1 & 90.1 & 92.1 & 24$^*$ \\ 
STCN~\cite{STCN} & $\times$ & 85.6 & 82.5 & 88.7 & 20$^*$ & 91.6 & 90.7 & 92.5 & 22$^*$ \\ 
\midrule 
FAVOS \cite{FAVOS} & \checkmark & 58.2 & 54.6 & 61.8 & \textless1 & 80.9 & 82.4 & 79.5 & \textless1 \\
FTMU~\cite{FTMU} & \checkmark & 70.6 & 69.1 & - & 11.1 & 78.9 & 77.5 & - & 11.1 \\
SAT~\cite{SAT} & \checkmark & 72.3 & 68.6 & 76.0 & \textless39 & 83.1 & 82.6 & 83.6 & 39 \\
TAN-DTTM~\cite{TAN_DTTM} & \checkmark & 75.9 & 72.3 & 79.4 & 7.1 & - & - & - & - \\
RMNet~\cite{RMNet} & \checkmark & 83.5 & 81.0 & 86.0 & \textless11.9 & 88.8  & 88.9  & 88.7 & 11.9 \\
DMN-AOA~\cite{DMN-AOA} & \checkmark & 84.0 & 81.0 & 87.0 & 6.3 & - & - & - & - \\
\textbf{RAVOS (Ours)} & \checkmark & \textbf{86.1} & \textbf{82.9} & \textbf{89.3} & \textbf{42} & \textbf{91.7} & \textbf{90.8} & \textbf{92.6} & \textbf{58} \\
\bottomrule 
\end{tabular}
\end{table*}

\subsection{Occluded Video Object Segmentation Dataset}
\label{subsec:ovos}
In this work, we introduce occluded video object segmentation (OVOS) dataset to evaluate the performance of VOS models under occlusions. OVOS is an extension of the training set of OVIS dataset~\cite{OVIS} in video instance segmentation since the segmentation of the first frame is not available for the validation set. 
To meet the format of DAVIS for convenient evaluation, we only select the objects that appear in the first frame as targets and resize videos to make their shortest size 480 pixels. An example is shown in Fig.~\ref{fig:ovos}, OVOS comes with accurate annotations and includes severe object occlusions. 
The presented OVOS dataset contains 607 video sequences with a total of 42149 frames and 2034 objects, which is larger than the current largest YouTube-VOS 2019 validation set (507 videos with a total of 13710 frames).
The dataset is available at \url{http://ieee-dataport.org/9608}.

\section{Implementation Details}
 
\textbf{Training.} Exactly following previous works~\cite{STCN,MIVOS}, we pre-train the models on static image datasets~\cite{DUTS,ECSSD,HRSOD,BIG,FSS1000} and perform the main training on the synthetic dataset~\cite{MIVOS} as well as DAVIS~\cite{DAVIS} and YouTube-VOS~\cite{YOUTUBE}. In the former stage, three synthetic frames are generated from one static image by applying random augmentation. During main training, all the video frames are resized to 480p, and three neighboring frames are randomly sampled with the maximum sampling interval ranging from 5 to 25.

In this work, all experiments are conducted in PyTorch~\cite{PyTorch} using a single 3090 GPU. Adam optimizer~\cite{Adam} is used to optimize the parameters. We adopt  bootstrapped cross-entropy loss $\mathcal{L}_{seg}$ to train the segmentation model and mean squared error loss $\mathcal{L}_{track}$ to train the OMT:
\begin{equation}
\mathcal{L}_{seg} = -  \frac{1}{n} \sum_{i=1}^{n}  \sum_{j=1}^{c} y_{i,j} \log f_{j}(x_{i};\theta)
\label{eq5}
\end{equation}
\begin{equation}
\mathcal{L}_{track} = \frac{1}{n} \sum_{i=1}^{n} {(v_{i}-\hat{v}}_{i})^2, v \in [x_{1},y_{1},x_{2},y_{2}]
\label{eq6}
\end{equation}

\begin{table*}[h]
\centering
\caption{Evaluation on DAVIS 2017 test-dev split.} \label{tab:davis17test}
\begin{tabular}{ccccccccc}
\toprule 
& STM~\cite{STM} & CFBI~\cite{CFBI} & KMN~\cite{KMN} & RMNet~\cite{RMNet} & GIEL~\cite{GIEL} & R50-AOT~\cite{AOT} & STCN~\cite{STCN} & \textbf{RAVOS (Ours)}  \\
\midrule
\( \mathcal{J} \)\&\( \mathcal{F} \) & 72.2 & 75.0 & 77.2 & 75.0  & 75.2 & 79.6 & 79.9 & \textbf{80.8}   \\
\( \mathcal{J} \) & 69.3 & 71.4 & 74.1 & 71.9 & 72.0  & 75.9 & 76.3 & \textbf{77.1}  \\
\( \mathcal{F} \) & 75.2 & 78.7 & 80.3 & 78.1 & 78.3 & 83.3 & 83.5 & \textbf{84.5}    \\
\bottomrule 
\end{tabular}
\end{table*}

\textbf{Inference.} We segment all videos at 480p during inference. Unless specified otherwise, RAVOS updates the memory every three frames for DAVIS and five frames for YouTube-VOS. We adopt only the top 20 matches for feature propagation as in~\cite{STCN}. For video segmentation, we segment entire features on the second frame and start to track objects on the third frame using the positional information of objects in the two past frames. The minimum object ROI to feature area ratio is set to 0.2 to avoid object features being too small to decode. The object ROI is expanded to the whole image when OMT senses the disappearance of objects, and RAVOS performs regional segmentation again when the objects appear again in subsequent frames.

\section{Experiments}

We evaluate RAVOS on the popular DAVIS and YouTube-VOS benchmark datasets as well as our newly created OVOS dataset. Region Similarity \( \mathcal{J} \) (average IoU score between the segmentation and ground truth), Contour Accuracy \( \mathcal{F} \) (average boundary similarity between the segmentation and ground truth), and their mean value \( \mathcal{J} \)\&\( \mathcal{F} \) are used as the evaluation metrics. All results are evaluated using the official evaluation tools or servers and, unless specified otherwise, FPS is measured without automatic mixed precision.

\subsection{Quantitative Results}
\label{Quantitative}

\textbf{DAVIS 2016}~\cite{DAVIS2016} is a popular single-object benchmark dataset that contains 20 videos in the validation set. For a fair comparison, we re-time the FPS for the nearest competitors on our machine, \ie, STM~\cite{STM}, R50-AOT~\cite{AOT}, and STCN~\cite{STCN}.
As shown in Table~\ref{tab:davis}, RAVOS achieves 91.7 \( \mathcal{J} \)\&\( \mathcal{F} \) with 58 FPS on DAVIS 2016 validation set, surpassing the above competitors in both accuracy and inference speed. RAVOS even outperforms STCN with significant faster inference speed (2.6$\times$ faster).

\textbf{DAVIS 2017}~\cite{DAVIS} is a multi-object extension of DAVIS 2016, which contains 30 videos in the validation and test-dev split, separately. As shown in Table~\ref{tab:davis}, although the proposed RAVOS aims at reducing redundant segmentation and memorization, it achieves 86.1 \( \mathcal{J} \)\&\( \mathcal{F} \) with 42 FPS, leading all present methods. Compared with the nearest regional segmentation competitor DMN-AOA, RAVOS achieves better performance (86.1 \vs{} 84.0 \( \mathcal{J} \)\&\( \mathcal{F} \)) and runs more than 5$\times$ faster. Compared with the nearest competitor STCN, RAVOS surpasses it by 0.5\% and runs about 2.1$\times$ faster (42 \vs{} 20 FPS). We further evaluate our method on DAVIS 2017 test-dev split. As shown in Table~\ref{tab:davis17test}, RAVOS achieves 80.8 \( \mathcal{J} \)\&\( \mathcal{F} \) and outperforms current state-of-the-art STCN by 0.9\%.

\begin{table}[t]
\setlength{\tabcolsep}{5pt}
% \footnotesize
\centering
\caption{Quantitative evaluation on YouTube-VOS 2018 and 2019 validation sets. $\mathcal{J}\&\mathcal{F}$ is the overall performance on ``seen" and ``unseen" categories. $RS$: regional segmentation. $^*$: re-timed on our machine for fair-comparison.} 
\label{tab:ytb}
\begin{tabular}{lccccccc}
\toprule 
 &  & &  \multicolumn{2}{c}{Seen}  & \multicolumn{2}{c}{Unseen} &  \\
\midrule 
Methods & RS & $\mathcal{J}\&\mathcal{F}$ & $\mathcal{J}$ & $\mathcal{F}$ & $\mathcal{J}$ & $\mathcal{F}$ & FPS \\
\midrule 
\multicolumn{7}{c}{\textit{Validation 2018 Split}} \\
\midrule 
STM~\cite{STM} & $\times$  & 79.4 & 79.7 & 84.2 & 72.8 & 80.9 & - \\
GC~\cite{GC} & $\times$  & 73.2 & 72.6 & 68.9 & 75.6 & 75.7 & - \\
GraphMem~\cite{GraphMem} & $\times$  & 80.2 & 80.7 & 85.1 & 74.0 & 80.9 & - \\
LWL~\cite{LWL} & $\times$  & 81.5 & 80.4 & 84.9 & 76.4 & 84.4 & - \\
KMN~\cite{KMN} & $\times$  & 81.4 & 81.4 & 85.6 & 75.3 & 83.3 & - \\
CFBI~\cite{CFBI} & $\times$  & 81.4 & 81.1 & 85.8 & 75.3 & 83.4 & 3.4 \\
AFB-URR~\cite{AFB_URR} & $\times$  & 79.6 & 78.8 & 83.1 & 74.1 & 82.6 & - \\
CFBI+~\cite{CFBI+} & $\times$  & 82.8 & 81.8 & 86.6 & 77.1 & 85.6 & 4.0 \\
SwiftNet~\cite{SwiftNet} & $\times$  & 77.8 & 77.8 & 81.8 & 72.3 & 79.5 & -\\
SST~\cite{SSTVOS} & $\times$  & 81.7 & 81.2 & - & 76.0 & - & - \\
GIEL~\cite{GIEL} & $\times$  & 80.6 & 80.7 & 85.0 & 75.0 & 81.9 & -\\
ASRF~\cite{ASRF} & $\times$  & 81.9 & 81.0 & 85.8 & 76.3 & 84.3 & -\\
LCM~\cite{LCM} & $\times$  & 82.0 & 82.2 & 86.7 & 75.7 & 83.4 & - \\
HMMN~\cite{HMMN} & $\times$  & 82.6 & 82.1 &  87.0 & 76.8 & 84.6 & - \\
R50-AOT~\cite{AOT} & $\times$  & 84.1 & \textbf{83.7} & \textbf{88.5} & 78.1 & 86.1 & 14.9 \\
STCN~\cite{STCN} & $\times$  & 84.3 & 83.2 & 87.9 & 79.0 & 87.3 & 12$^*$ \\ 
\midrule 
SAT\cite{SAT} & \checkmark & 63.6 & 67.1 & 70.2 & 55.3 & 61.7 & - \\
TAN-DTTM~\cite{TAN_DTTM} & \checkmark & 73.5 & - & - & - & - & - \\
RMNet~\cite{RMNet} & \checkmark & 81.5 & 82.1 & 85.7 & 75.7 & 82.4 & -\\
DMN-AOA~\cite{DMN-AOA} & \checkmark & 82.5 & 82.5 & 86.9 & 76.2 & 84.2 & - \\
\textbf{RAVOS (Ours)} & \checkmark & \textbf{84.4} & 83.1 & 87.8 & \textbf{79.1} & \textbf{87.4} &  \textbf{23} \\
\midrule 
\multicolumn{7}{c}{\textit{Validation 2019 Split}} \\
\midrule 
STM~\cite{STM} & $\times$  & 79.3 & 79.8 & 83.8 & 73.0 & 80.5 & - \\
CFBI~\cite{CFBI} & $\times$  & 81.0 & 80.6 & 85.1 & 75.2 & 83.0 & 3.4 \\
CFBI+~\cite{CFBI+} & $\times$  & 82.6 & 81.7 & 86.2 & 77.1 & 85.2 & 4.0 \\
SST~\cite{SSTVOS} & $\times$  & 81.8 & 80.9 & - & 76.6 & - & - \\
HMMN~\cite{HMMN} & $\times$  & 82.5 & 81.7 & 77.3 & 86.1 & 85.0 & - \\
R50-AOT~\cite{AOT} & $\times$  & 84.1 & \textbf{83.5} & \textbf{88.1} & 78.4 & 86.3 & 14.9 \\
STCN~\cite{STCN} & $\times$  & \textbf{84.2} & 82.6 & 87.0 & \textbf{79.4} & \textbf{87.7} & 11$^*$ \\
\midrule
\textbf{RAVOS (Ours)} & \checkmark & \textbf{84.2} & 82.6 & 87.0 & \textbf{79.4} & \textbf{87.7} & \textbf{20} \\
\bottomrule 
\end{tabular}
\end{table}

\begin{table}[t]
% \setlength{\intextsep}{-10pt}
% if overlapped with main content, change [] below
% \begin{wraptable}[8]{r}{0.47\textwidth}
\setlength{\belowcaptionskip}{0.3cm}
% \footnotesize
\centering
\caption{Evaluation on OVOS dataset.} \label{tab:ovis}
\setlength{\tabcolsep}{4pt}
\begin{tabular}{lcccc}
\toprule 
Method & \( \mathcal{J} \)\&\( \mathcal{F} \) & \( \mathcal{J} \) & \( \mathcal{F} \) & FPS\\
\midrule
STCN  & 61.5 & 57.3 & 65.6 & 5.7 \\ 
\textbf{RAVOS (Ours)} & \textbf{62.5} & \textbf{58.3} & \textbf{66.6} & \textbf{14}\\ 
\bottomrule 
\end{tabular}
\end{table}

\begin{figure*}[t!]
\centering
\includegraphics[width=0.97\linewidth]{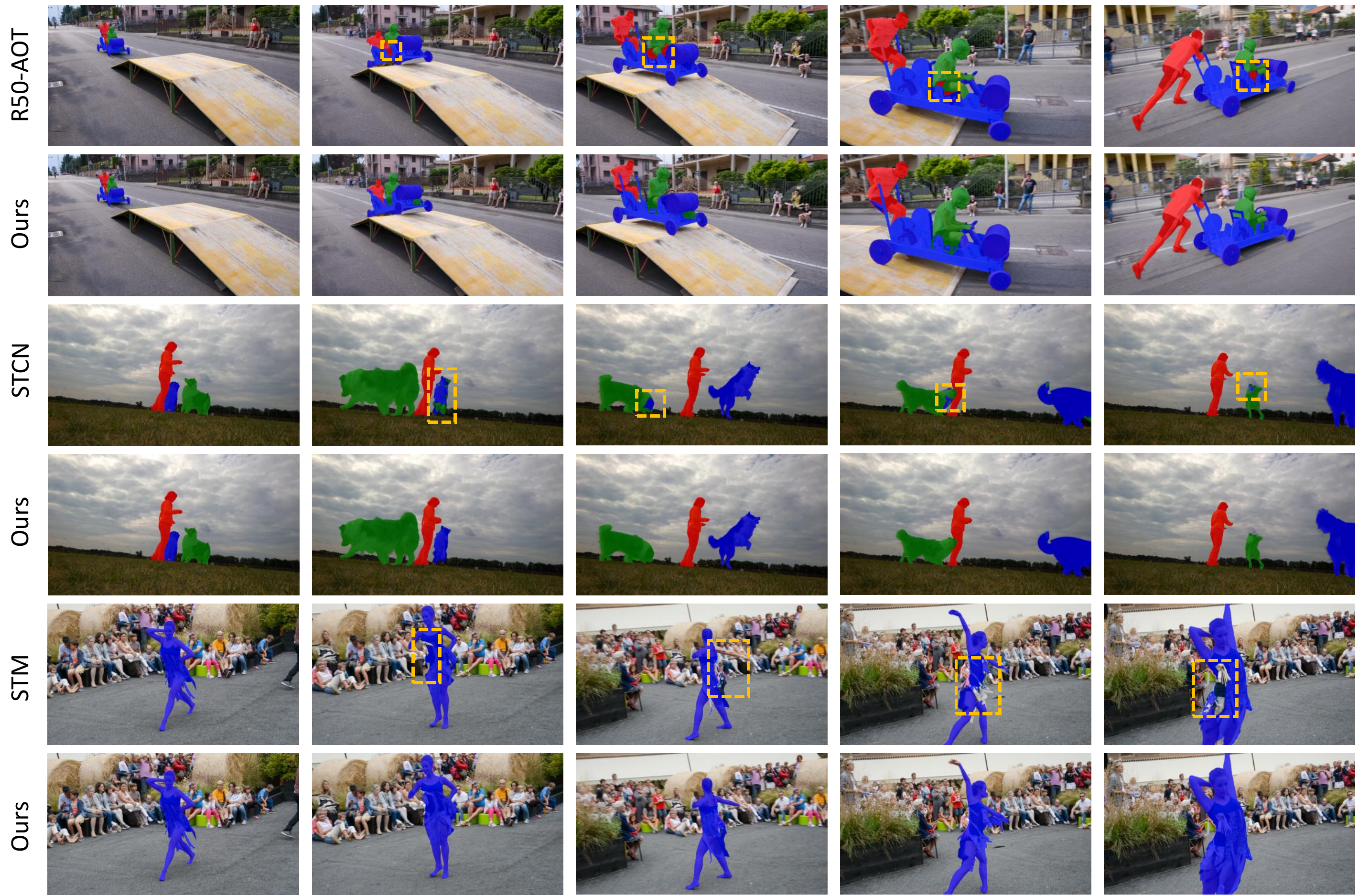}
\caption{Qualitative results. The first column denotes the reference frames for mask propagation.}
\label{fig:qualitative}
\end{figure*}

\begin{table*}[t!]
\centering
\caption{Ablation of motion path memory (M) and object-level segmentation (O) on DAVIS 2017 validation set.} \label{tab:abmodule} 
\begin{tabular}{llccccccc}
\toprule 
M & O &  \( \mathcal{J} \)\&\( \mathcal{F} \) &  \( \mathcal{J} \) & \( \mathcal{F} \) & Matching (ms) & Decoding (ms) & Mem. Size (K) & FPS \\
\midrule 
$\times$ & $\times$ & 85.6 & 82.5 & 88.7 & 12.2 & 7.5 & 36.3 & 20 \\ 
\checkmark  & $\times$ & 85.4 & 82.3 & 88.5 & 3.2 & 7.1 & 19.2 & 30 \\ 
$\times$ & \checkmark & 86.1 & 82.9 & 89.3 & 5.9 & 3.9 & 36.3 & 31 \\ 
\checkmark & \checkmark & \textbf{86.1} & \textbf{82.9} & \textbf{89.3} &  \textbf{2.2} &  \textbf{3.8} & \textbf{18.6} & \textbf{42} \\
\bottomrule 
\end{tabular}
\end{table*}

\textbf{YouTube-VOS}~\cite{YOUTUBE} is currently the largest dataset for VOS, containing 3471 videos in the training set and 474/507 videos in the 2018/2019 validation set. YouTube-VOS splits the validation videos into seen categories and unseen categories based on whether the objects of a category appear in the training videos or not. The performance on unseen categories is used to evaluate the generalization ability of models. As shown in Table~\ref{tab:ytb}, RAVOS achieves 84.4/84.2 \( \mathcal{J} \)\&\( \mathcal{F} \) and 23/20 FPS on YouTube-VOS 2018/2019 validation set. Compared with the nearest regional segmentation competitor DMN-AOA, RAVOS outperforms it by 1.9\%. Compared with the nearest competitor STCN, RAVOS has competitive performance and runs about 2$\times$ faster (23/20 \vs{} 12/11 FPS on YouTube-VOS 2018/2019). Overall, RAVOS achieves state-of-the-art performance with the fastest inference time.

\textbf{OVOS} is a large-scale occluded VOS dataset that contains 607 video sequences with severe object occlusions for validation. More details of the dataset are included in Section~\ref{subsec:ovos}. We directly evaluate RAVOS on OVOS dataset without retraining to verify its performance in the occlusion scenario. It is noteworthy that OVOS dataset is only used for evaluation and, to the best of our knowledge, this is the first time a semi-supervised VOS method is evaluated on this large-scale dataset. We also evaluate the state-of-the-art STCN on OVOS for comparison. Automatic mixed precision is used for both methods since OVOS contains some long video sequences, which cause large memory burden and out-of-memory problems for STCN. As shown in Table~\ref{tab:ovis}, RAVOS outperforms STCN by 1.0\% since our method performs regional segmentation to reduce the risk of false positives caused by same class object occlusions. More importantly, thanks to the efficient regional segmentation and memorization approaches, our method runs about 2.5$\times$ faster than STCN. The results also indicate that precisely localizing and reasoning under occlusions is still challenging for existing VOS models.

\subsection{Qualitative Results}

Fig.~\ref{fig:qualitative} shows qualitative results of RAVOS compared with STM, R50-AOT, and STCN. RAVOS performs better when multiple objects overlap with each other because it only segments the region within ROIs for each object. This reduces the risk of false positives on redundant context.

\begin{figure*}[t]
\centering
\includegraphics[width=0.95\linewidth]{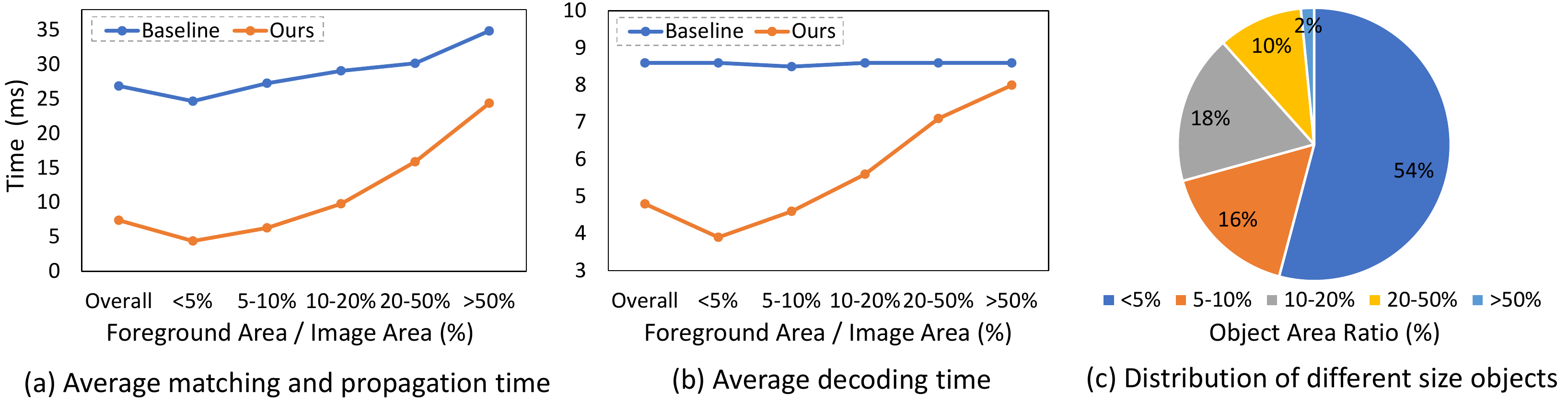}
\caption{Processing time for different object sizes on YouTube-VOS 2018 validation set. (a) Average spatio-temporal feature matching and propagation time (ms) on different object sizes; (b) Average feature decoding time (ms) on different object sizes; (c) Distribution of object area ratio.} 
\label{fig:statistic}
\end{figure*}

\begin{table*}[t]
\centering
\begin{minipage}{.3\textwidth}
\centering
\caption{RAVOS performance using different trackers.} \label{tab:abtracker}
\begin{tabular}{lcc}
\toprule
Tracker &  \( \mathcal{J} \)\&\( \mathcal{F} \) &  Time (ms) \\
\midrule 
Lucas–Kanade~\cite{LUCAS} & 80.0 & 217.4 \\ 
RAFT~\cite{RAFT} & 86.0 & 20.3\\ 
OMT (Ours) & \textbf{86.1} & \textbf{0.2}\\
\bottomrule 
\end{tabular}
\end{minipage}
\hfill
\begin{minipage}{.3\textwidth}
\centering
% \footnotesize
\caption{RAVOS performance using different memory regions.} \label{tab:memregion}
\begin{tabular}{lccc}
\toprule 
Memory &  \( \mathcal{J} \)\&\( \mathcal{F} \) &  \( \mathcal{J} \) &  \( \mathcal{F} \)  \\
\midrule 
Foreground & 85.3 & 82.1 & 88.5  \\ 
Motion Path & \textbf{86.1} & \textbf{82.9} & \textbf{89.3} \\ 
\bottomrule 
\end{tabular}
\end{minipage}
\hfill
\begin{minipage}{.3\textwidth}
\centering
% \footnotesize
\caption{RAVOS performance using different motion functions.} \label{tab:aborder}
\begin{tabular}{lcccc}
\toprule 
Function &  \( \mathcal{J} \)\&\( \mathcal{F} \) &  \( \mathcal{J} \) &  \( \mathcal{F} \) \\
\midrule 
Linear & 86.0 & 82.8 & \textbf{89.3} \\
Quadratic & \textbf{86.1} & \textbf{82.9} & \textbf{89.3} \\
Cubic & 86.0 & 82.8 & 89.2 \\
\bottomrule 
\end{tabular}
\end{minipage}
\end{table*}

\subsection{Ablation Studies}

\textbf{Object-level segmentation with object motion tracker.}
Table~\ref{tab:abmodule} shows the ablation study for object-level segmentation. RAVOS outperforms baseline by 0.5\% by segmenting the predicted object ROIs because of the reduced risk of false positives on background regions. Moreover, object-level segmentation significantly accelerates the feature matching time (5.9 \vs{} 12.2 ms) as well as feature decoding time (3.9 \vs{} 7.5 ms) due to the less redundant computations.

\textbf{Motion path memory.}
As shown in Table~\ref{tab:abmodule}, by memorizing the important motion path regions instead of the entire features, \( \mathcal{J} \)\&\( \mathcal{F} \) drops by 0.2 on DAVIS 2017 validation set. This is because MPM filters out most of the background regions far from objects before updating the memory, resulting in the loss of some prior information of backgrounds when performing global segmentation. However, leveraging MPM for object-level segmentation does not drop the performance of the model since object-level segmentation avoids segmenting these redundant background areas. Most importantly, MPM significantly reduces the feature matching time by about 3.8$\times$ (3.2 \vs{} 12.2 ms) and memory size by about 1.9$\times$ (19.2 \vs{} 36.3 K). This is important when segmenting long video sequences, \ie, when the method is deployed on autonomous systems. In a nutshell, OMT reduces redundant segmentation to accelerate object segmentation, and MPM reduces redundant memorization to accelerate spatio-temporal feature matching and propagation. OMT and MPM are complementary, and the combination of the two modules achieves the best performance.

\textbf{Different trackers.}
We compare the performance of OMT with the traditional Lucas-Kanade optical flow~\cite{LUCAS} and the cutting-edge RAFT optical flow~\cite{RAFT}. As shown in Table~\ref{tab:abtracker}, OMT slightly outperforms the two methods in regional VOS. Most importantly, OMT only requires 0.2 ms for single object tracking, which is about 100$\times$ faster than RAFT. These results indicate that OMT is more suitable for object tracking in efficient VOS.

\textbf{Different memory regions.} We compare our MPM with foreground bounding boxes only memory, which does not include the motion path. As shown in Table~\ref{tab:memregion}, MPM outperforms foreground bounding boxes only memory by 0.8\% since MPM also contains some useful background features for the ROIs in next frame.

\textbf{Different tracking functions.} We use different motion functions to verify the performance of OMT. As shown in Table~\ref{tab:aborder}, all motion functions have good performance and the quadratic motion function obtains the best performance among the three. The results indicate the efficacy of the motion estimator in OMT in predicting the parameters of motion functions.

\textbf{Inference time analysis.}
We first compute the distribution of object area ratio on YouTube-VOS 2018 validation set, where the size of an object is determined by the given mask of the first frame. As shown in Fig.~\ref{fig:statistic} (c), nearly 70\% of objects are smaller than 10\% of the image area in YouTube-VOS 2018 validation set. We then analyze the single object processing time of feature matching and propagation as well as feature decoding to observe the efficiency of our method on different object sizes. As shown in Fig.~\ref{fig:statistic}  (a) and (b), RAVOS significantly accelerates the feature matching and propagation as well as feature decoding time on small objects. That is because the smaller the objects, the smaller is their ROIs, and the faster our method executes.

\section{Conclusion}
In this paper, we presented a novel segmentation-by-tracking approach for region aware semi-supervised VOS. Our method outperformed existing techniques on multiple benchmark datasets in accuracy with the added advantage of faster inference time. We proposed OMT which meets the requirements of fast processing and minimal redundancy to achieve a very high frame rate of 5000 FPS for object tracking and ROI prediction. On top of OMT, we designed object-level segmentation and MPM to accelerate VOS and reduce memory size by a large margin. Moreover, we evaluated RAVOS on a newly created OVOS dataset for the first time in the community of semi-supervised VOS. We hope our RAVOS can serve as a fundamental baseline for efficient VOS and help in the advancement of research and deployment of efficient video object segmentation, video instance segmentation, and multiple object tracking.

% or


% \section*{Acknowledgment}


\ifCLASSOPTIONcaptionsoff
  \newpage
\fi



% \newpage
\bibliographystyle{IEEEtran}
\bibliography{egbib}


\vfill
%\newpage


%\vfill


\end{document}